# PointNeXt Adapter Improved Model for BIM Component Defect Association under Construction Images and Design Model Attributes

Na Yu [a]

[a] First Highway Engineering Co., Ltd. Chengdu 610000, Sichuan , China

**Abstract**

Construction site defect analysis increasingly requires linking field evidence to the correct BIM object, because a crack, spalling area, missing sleeve or misaligned support is useful for quality control only when it is connected with component identity, location, responsibility and design intent. This paper proposes a PointNeXt Adapter improved model for BIM component defect association under construction images and design model attributes. The method aligns site image crops, local point cloud patches, BIM component identifiers, geometry attributes, material categories, installation stage, inspection distance, viewpoint quality and source reliability tags into component level inspection windows. PointNeXt learns local geometry and deviation from as built point evidence, while the adapter injects BIM attributes so the same geometric irregularity can be interpreted differently for structural, envelope, MEP and embedded components. The fusion head outputs component link, defect class, defect probability, confidence, dominant evidence and recommended review action. A synthetic multisource BIM inspection dataset is used to evaluate repeated layouts, occlusion, sparse point clouds, registration errors and unseen component arrangements. Results show that attribute aware point learning improves association accuracy, defect recall and confidence calibration compared with image only, point only, simple fusion and adapter free baselines. The framework is positioned as traceable BIM issue review support, requiring manual confirmation and staged field validation before operational deployment.

## 1. Introduction

BIM-assisted construction inspection requires linking site observations to the correct design component because a visible defect is useful only when it can be connected with location, component identity and design responsibility [1], and this requirement makes defect association different from general image classification. Construction-image analytics can identify cracks, spalling, exposed reinforcement and installation errors [2], but the same visual cue may belong to different components under different viewpoints. Point-cloud inspection can measure geometric deviation and missing elements [3], yet point evidence alone may not identify material condition or surface damage in cluttered scenes. BIM attribute integration has therefore become important for model-based quality review [4], and the present paper follows this direction by combining images, point clouds and design attributes in one component-level prediction window. The goal is to determine not only whether a defect is present, but also which BIM component it should be associated with and how reliable that association is.

Traditional site inspection compares photographs, drawings, notes and BIM models manually. This workflow is understandable, but it becomes slow and inconsistent when many similar components appear in the same area. Surface cracks, displaced pipe brackets, damaged embedded plates and missing sleeves can be visually small, yet their engineering meaning depends on the linked component. A purely visual model may treat a temporary shadow as a defect, while a point-cloud model may detect deviation without locating material damage. The inspection model should therefore retain component semantics, design attributes and measured geometry in one decision window.

This paper develops a PointNeXt-Adapter improved model for BIM component defect association. Section 2 introduces BIM component inspection, construction-site visual evidence, point-cloud learning and attribute-aware fusion. Section 3 introduces image-BIM-point cloud window encoding and the PointNeXt-Adapter association network. Section 4 is the simulated BIM component dataset, the baseline comparison, adapter contribution, occlusion robustness and engineering review output. Section 5 concludes the study and presents deployment-oriented extensions of BIM-based quality management.

## 2. Related Work

### 2.1. BIM Component Defect Association and Construction-site Visual Evidence

BIM has been applied to organise construction data, component attributes, spatial locations and inspection records. In quality inspection, BIM helps link field evidence with the identity and design requirements of components [5], and defect association is therefore more precise than in other types of defect recognition. A visible crack or other damage to the surface of a wall, slab, beam, pipe support or embedded part needs to be reported to the engineer for repair. Thus, a problem of semantic matching between site observations and design-model entities has arisen.

Construction-site pictures are direct evidence, but they are subject to occlusion, temporary equipment, changes in light and repeated component patterns. Although image-based defect detection has been improved [6], the problem of component association in overlapping objects within an image crop remains. For example, there may be a defect in the pipe near the wall opening, and a damaged bracket may be partially concealed by temporary cables. Therefore, the model needs to be connected with visual information and BIM data, such as component type, Design location, material and installation stage.

BIM-based inspections should also correspond to design-as-built. Design attributes generally show the desired shape, size and connection of components, and construction progress may be irregular. A component has been partially installed, is covered with formwork, or displaced from its original position. Research on scan-to-BIM and progress monitoring [7] shows that geometric matching can be used when visual evidence is insufficient. Point cloud features are employed to offer as-built geometry, and image evidence is used to provide surface and material clues.

The association task is also logged. A defect prediction without a component link cannot be added to the BIM issue list easily, and a component link without confidence may be misleading to quality management. Most research on construction digitalisation has focused on traceable review records [8], so the output of the model should include component identity, defect probability, confidence and dominant evidence. Currently, a review leader has been appointed, and the three types of decisions are image-driven, point-driven, and attribute-driven.

### 2.2. Point-cloud Learning and Attribute-aware Fusion for Built-environment Inspection

Point-cloud learning provides an inherent representation of local geometry, shape variation and neighbourhood for regions. PointNet-style models were the first to learn directly from unordered point sets, and subsequently, architectures have introduced local geometric information via neighbourhood aggregation [9]. PointNeXt further improves the encoding of local structure and is suitable for building inspection; construction defects are often in the form of local deformation, missing surfaces or abnormal offsets. Point features of the BIM component review can show whether the site component meets the design geometry.

Point geometry is incapable of connecting defects at the component level. The two parts have the same shape but different functions, materials or tolerances. Therefore, attribute-aware fusion has been gradually applied in the analysis of digital construction [10], and the same geometric deviation may be acceptable in some places but not in others. The design of the adapter in this paper injects BIM component attributes into the point representation, and thus can read geometry through the lens of design semantics.

Fusion models that combine image, point and attribute evidence still need source reliability. A camera crop may be blurred, a point cloud may be sparse, and BIM attributes may be outdated if field changes have not been updated. Vassilev et al. [11] showed that uncertainty-aware point-cloud segmentation improves infrastructure-project analysis by representing unreliable geometric evidence instead of treating all points as equally dependable. The revised model therefore keeps reliability labels for each source, so a low-confidence association is routed to manual review rather than directly becoming a BIM issue.

Explainability is also required for construction quality review. The manager should know whether a warning is caused by visible damage, geometric mismatch, design-attribute conflict or weak source quality. Liang et al. [12] showed that explainable vision-based inspection remains more reliable under visual interference when the decision path is linked to interpretable evidence. The current model therefore reports component association, defect probability, dominant evidence and confidence together, allowing the linked component to be checked against the evidence.

## 3. Methodology

### 3.1. Image-BIM-Point Cloud Window Encoding and Component Matching

First, build a component-level inspection window to achieve this. Each window shows a cropped image of the site, a local point cloud segment, BIM component attributes, camera-view metadata, point-cloud registration indicators and inspection-stage tags. A candidate BIM component is used to model the window, and thus, visual and geometric information can be linked to the same design element. Therefore, the structure is used to prevent the problem of subjective defect association. It is a matching problem of observed evidence, as-built geometry and as-designed component semantics.

Figure 1 is the component-window encoding framework. The left side contains a site image crop, local point-cloud patch and BIM attribute card for a candidate component. The center aligns these sources through component ID, spatial registration and inspection time, while reliability tags record occlusion, point density, noise and registration error. The right side forms a component-level inspection window that preserves image evidence, point geometry, design attributes and source reliability for defect association.

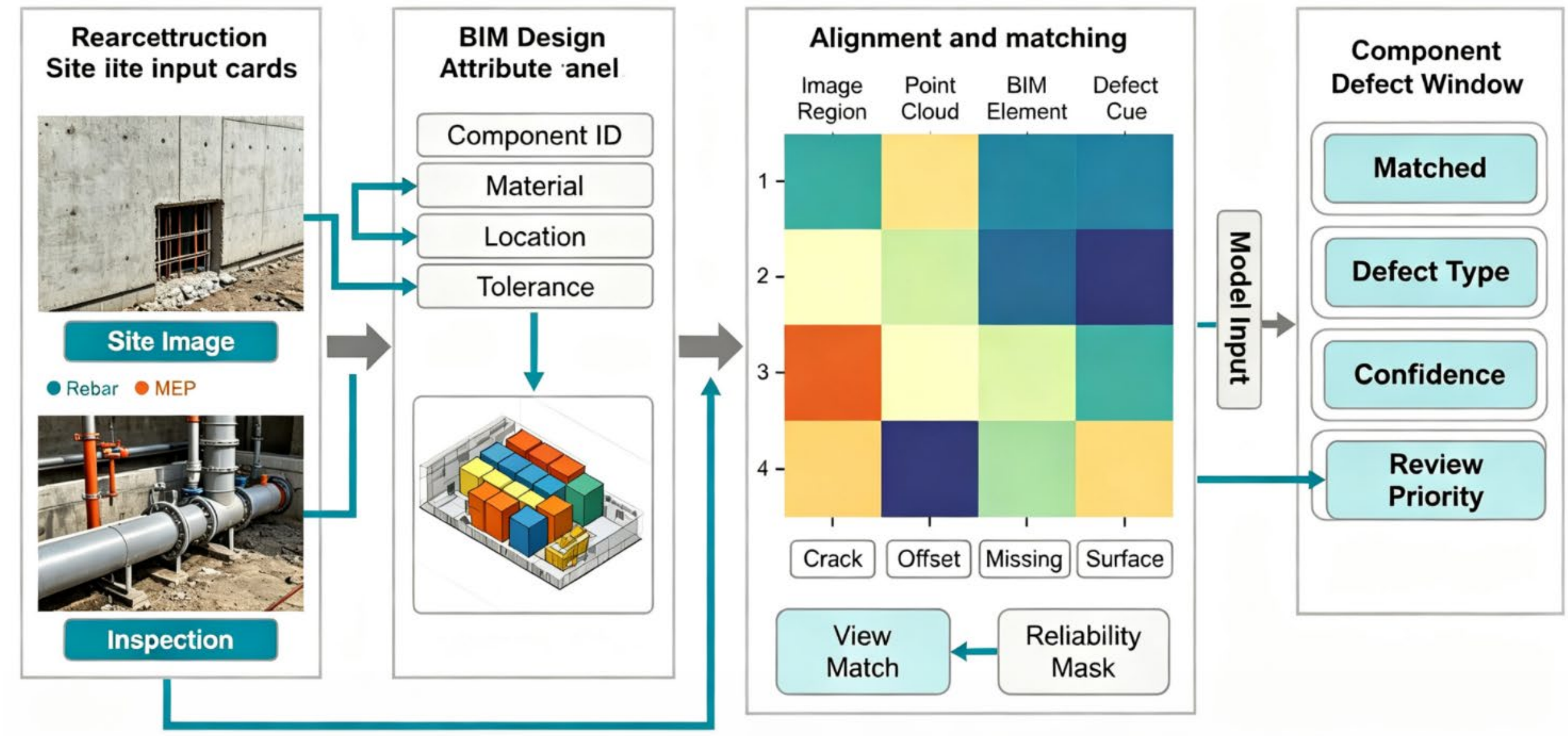

**Figure 1. Image-BIM-point cloud component-window encoding framework.**

$$Wc = \{Ic, Pc^{a}, Ac, Rc\} \tag{1}$$

Here Wc denotes the inspection window for component c. Ic is the image evidence, Pc is the local point-cloud evidence, Ac is the BIM attribute vector, and Rc is the source-reliability vector. Keeping these groups separate helps the model identify whether an association is supported by visual evidence, geometric evidence or design attributes.

$$Ac = [tc, mc, lc, gc, sc] \tag{2}$$

The attribute vector Ac includes component type, material, designed location, geometry class and construction stage. These attributes provide design context so that similar visual defects can be interpreted differently for structural, envelope, MEP and embedded components.

$$Dc = NN(Pc, Bc) \tag{3}$$

The deviation descriptor Dc is obtained by matching the local point cloud Pc to the BIM geometry Bc. It records as-built and as-designed differences without requiring the entire site model to be processed at once.

$$rs = 1 - clip(os + ns + es, 0,1) \tag{4}$$

The reliability score r s combines occlusion, noise and registration-error indicators for source s. A weak source is not removed, but its contribution is reduced during fusion so that the model remains cautious under partial visibility.

### 3.2. PointNeXt-Adapter Defect Association Network

The new network has an image-evidence branch, a PointNeXt geometry branch, a BIM attribute adapter, and a fusion review head. The image branch extracts visible damage clues from the site photos. The PointNeXt branch learns local point geometry and the displacement of the design model. Add component type, material, geometry class, and construction stage to the point information in the adapter. The fusion head outputs component association, defect probability, confidence, dominant evidence and review action.

Figure 2 is the model architecture. The lower level contains image, point-cloud and BIM-attribute encoders. The middle layer is the PointNeXt-Adapter, which adjusts local geometric features according to design-model attributes rather than merely concatenating sources. The top level combines visual damage indicators, geometric irregularities and semantic attributes. The right side shows the association head and review head, which produce component links, defect status, confidence and dominant evidence for BIM issue management.

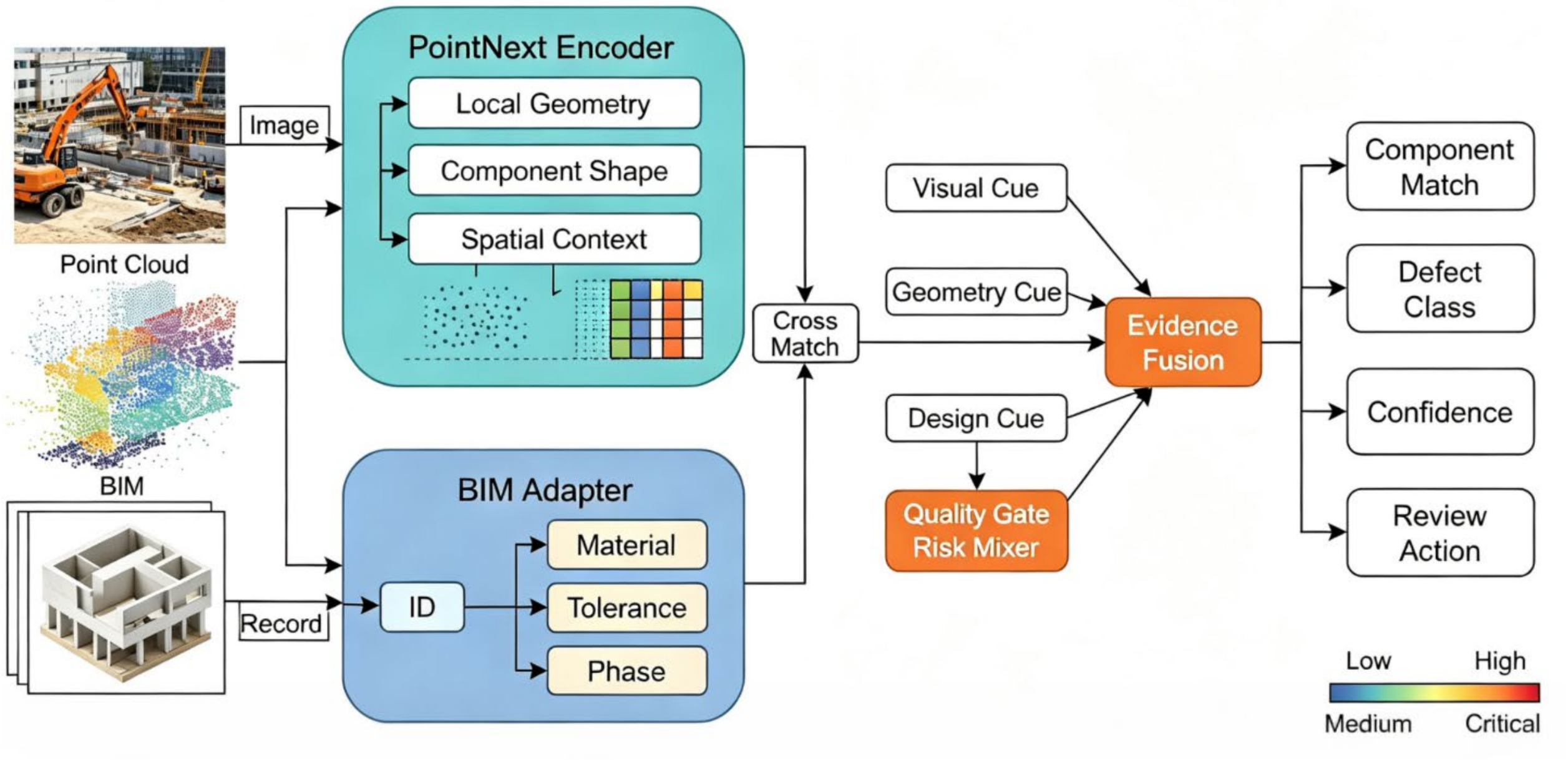

**Figure 2. PointNeXt-Adapter defect association network.**

$$Gc = \text{PointNeXt}\ (Pc, Dc) \tag{5}$$

The geometric representation Gc is learned from local point coordinates and deviation descriptors. PointNeXt is used because local neighbourhood aggregation can preserve small offsets, missing surfaces and spatial discontinuities around the component.

$$A'c = \text{Adapter}\ (Ac) \tag{6}$$

The adapted attribute feature A'c transforms design-model attributes into a feature space compatible with point geometry. This adapter helps the model interpret geometric deviation according to component type and construction stage.

$$Zc = \text{Fusion}(\ lc, Gc, A'c, Rc\ ) \tag{7}$$

The fused representation Zc combines image evidence, point geometry, adapted BIM attributes and reliability scores. Reliability weighting is included because occluded images and sparse point clouds should not produce the same confidence as complete inspection evidence.

$$pc = \text{SoftMax}\ (fc\ (Zc)) \tag{8}$$

The association probability pc estimates which BIM component should be linked with the observed defect evidence. It is interpreted together with confidence because repeated components may create ambiguous candidate matches.

$$L\ = \text{Lassoc} + \lambda 1\ \text{Ldef} + \lambda 2\ \text{Lconf} + \lambda 3\ \text{Lreg} \tag{9}$$

The total loss combines component association, defect classification, confidence calibration and registration consistency. This design encourages the model to learn a correct component link while preserving reliable defect evidence and stable geometry alignment.

## 4. Results and Discussion

### 4.1. Simulated BIM Component Dataset and Defect-label Distribution

A synthetic multi-source BIM inspection dataset is constructed for component-level defect association. It contains 18,000 component inspection windows, four component groups, five defect categories, three occlusion levels and four source-quality labels. The training, validation and test splits are scenario-aware, and unseen component layouts and occluded-view cases are included in the test set. BIM inspection studies support simulation-assisted evaluation when field defects are rare and project-specific [13], so this dataset is used as a controlled benchmark rather than proof of field readiness.

The Dataset has real-life factors and links to evidence. Beams, slabs, columns, shear walls, etc., are the components. Envelope components are panels, doors, windows, and the outer wall. MEP Components are Pipes, Ducts and Supports. Sleeves, Plates, and Anchors are the parts. Each inspection window has an image crop, a local point-cloud patch and BIM attributes. To link the evidence with the Design Features at the component level [14], the labels should contain both the defect status and other necessary component information. The above structure will determine whether the model can recognise a real component defect close to other objects or close to other defects.

Table 1 shows the numerical results of the model evaluation. The four groups of components have the following inspection window, defect ratio and mean matching IoU data in the table. The defect rate for the embedded-parts group is 0.45, and its mean matching IoU of 0.73 is relatively low; thus, small and partially hidden components are harder to match. Datasets of construction defects are often unbalanced [15], so before presenting the model results, a data table is provided to show the evaluation difficulty. As shown in the table, the experiments are not simple image classification problems, and the matching quality of each group is different. According to the data, a single global accuracy score is also not suitable. The matching IoU for the MEP and embedded-part groups is relatively low; therefore, their errors may be caused by association ambiguity rather than defect recognition errors. In the later performance interpretation, these groups will be selected for individual assessment due to a high defect rate and poor matching results. The data-table design of the experiment environment has not omitted the difficult-to-explain groups of components in the model comparison.

**Table 1. BIM component defect dataset statistics for model evaluation.**

| Component group | Windows | Defect ratio | Mean matching IoU |
|---|---|---|---|
| Structure | 6240 | 0.37 | 0.82 |
| Envelope | 4860 | 0.41 | 0.79 |
| MEP pipeline | 3920 | 0.34 | 0.76 |
| Embedded parts | 2980 | 0.45 | 0.73 |

Figure 3 is the simulated dataset and inspection-source distribution. As shown in Figure 3(a), the component-type panel separates structural, envelope, MEP and embedded components, so the model must associate evidence across different BIM entity types. Figure 3(b) includes crack, spalling, deformation, missing part and misalignment cases that cannot be resolved by one visual cue. As shown in Figure 3(c), image quality ranges from clear to blurred and occluded. As shown in Figure 3(d), point-cloud density ranges from dense to sparse. Scan-to-BIM inspection research shows that these source-quality differences affect defect association [16], so the panels justify reliability-aware fusion.

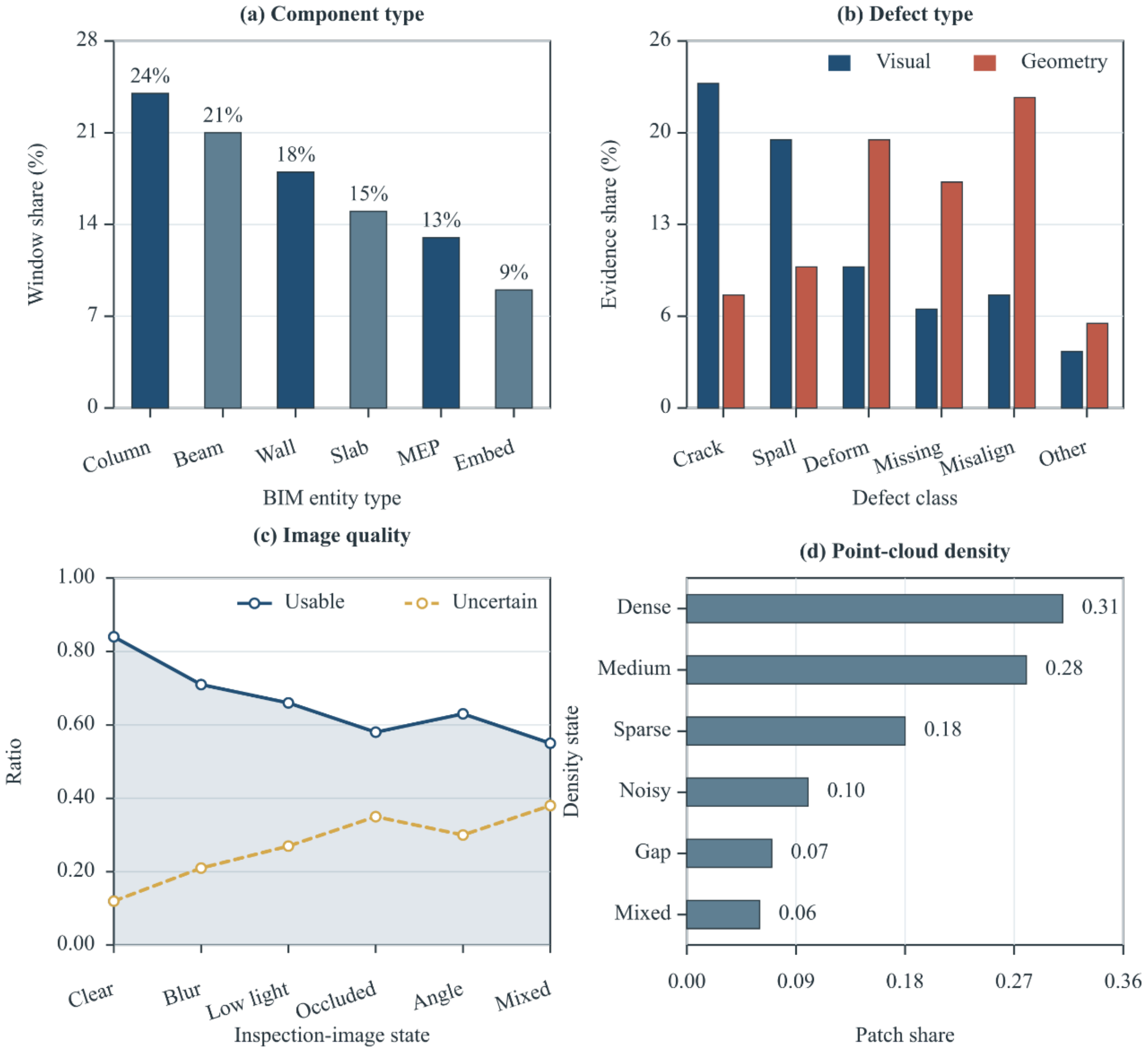

**Figure 3. BIM component dataset and inspection-source distribution. (a) component type. (b) defect type. (c) image quality. (d) point-cloud density.**

Figure 4 is the defect-label and matching-quality distribution. Figure 4(a) shows normal, minor defect, major defect and uncertain-review labels, and the uncertain-review class prevents weak evidence from being turned directly into a serious BIM issue. Figure 4(b) shows high, medium and low BIM-to-point alignment quality. Embedded and MEP components often have lower matching quality because they are smaller, partially hidden or repeated. The label design therefore connects defect severity with matching confidence before issue creation.

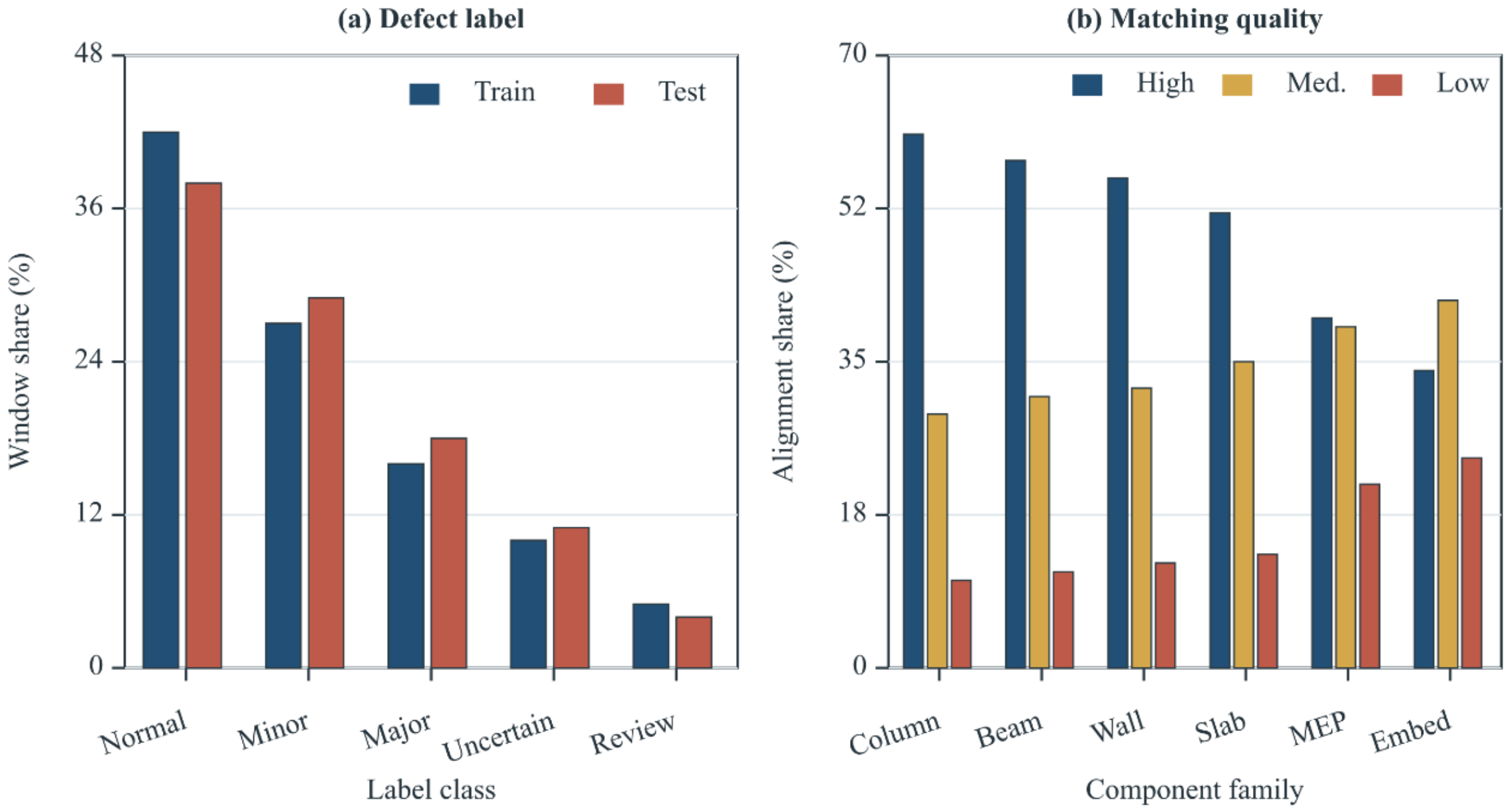

**Figure 4. Defect label and BIM matching-quality distribution. (a) defect label. (b) matching quality.**

### 4.2. Baseline Comparison and Attribute-adapter Contribution

The proposed PointNeXt-Adapter model is compared with an image-only CNN, a point-only PointNeXt model, a simple feature-concatenation Transformer, a PointNeXt model without BIM attributes and an adapter-free multi-source model. These comparisons test whether improvement comes from geometric learning, image evidence, design attributes or adapter-based fusion. Multi-source inspection studies show that visual, geometric and semantic evidence can improve construction quality assessment [17], and the current comparison indicates that the adapter helps the model use these sources coherently.

The image-only baseline is suitable for defects that are easily visible and clearly bounded. Performance Reduction Due to Occlusion and Redundant Modules. A point-only baseline can only find geometric defects and is unable to detect surface cracks or other damage. The overall accuracy of the combined model is relatively high, but some instances of false information are not detected. The adapter-free model is also less accurate in regions where modifications to the design attributes change the meaning of the same geometry. Attribute-aware BIM analytics have shown that design semantics can change the interpretation of defects [18], and thus, an adapter needs to be developed rather than a simple addition. It can help separate algorithm errors from data-collection errors in the subsequent project review.

Figure 5 compares defect association performance across models. Figure 5(a) reports association accuracy and shows whether site evidence is linked to the correct BIM component. Figure 5(b) reports defect recall, indicating whether real defects are missed. As shown in Figure 5(c), confidence calibration evaluates whether occluded or weakly registered cases are overconfident. Figure 5(d) reports review latency, showing whether the method remains practical for repeated site-inspection cycles. These metrics should be interpreted together rather than reduced to one accuracy value.

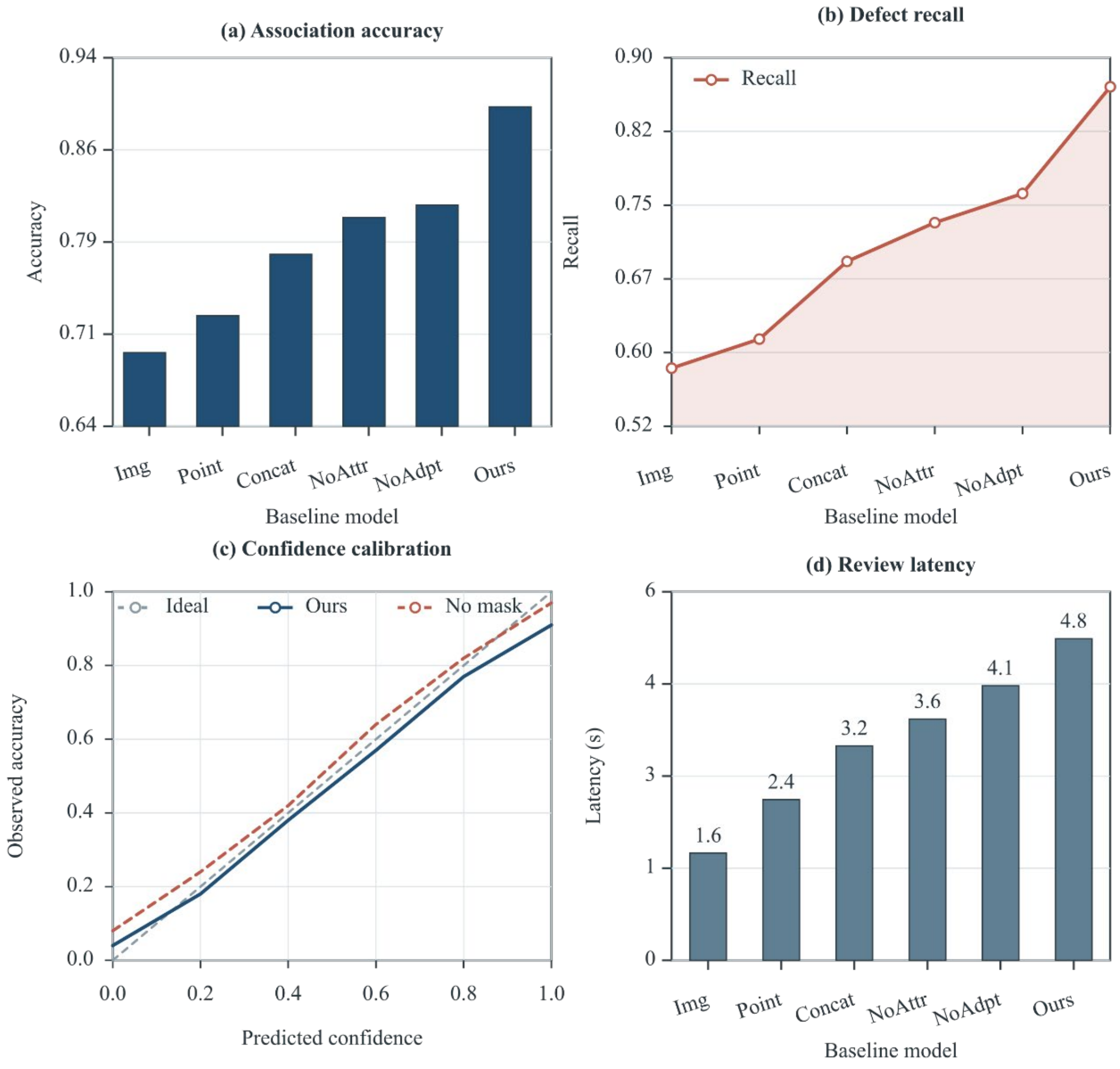

**Figure 5. Defect association performance under baseline comparison. (a) association accuracy. (b) defect recall. (c) confidence calibration. (d) review latency.**

Ablation studies clarify the role of each source. Removing BIM attributes makes component recognition harder because geometry must be interpreted without design semantics. Removing point-cloud evidence weakens misalignment and missing-part recognition, while removing images reduces crack and spalling detection. Without the adapter, cross-component generalization degrades because point features are not mapped into a BIM semantic context. Construction-automation research shows that different evidence sources have different failure modes [19], and the ablation results confirm this in the BIM component setting.

Figure 6 is the adapter and source-ablation sensitivity analysis. As shown in Figure 6(a), removing BIM attributes harms component association because semantic context is needed to choose the correct design element. Figure 6(b) shows that the image branch is critical for surface defects. As shown in Figure 6(c), the point branch is needed for deviation and missing-part cases. Figure 6(d) shows that removing the reliability mask makes confidence less stable under occlusion and registration uncertainty. The panels demonstrate that surface, geometry and attribute evidence correct different errors.

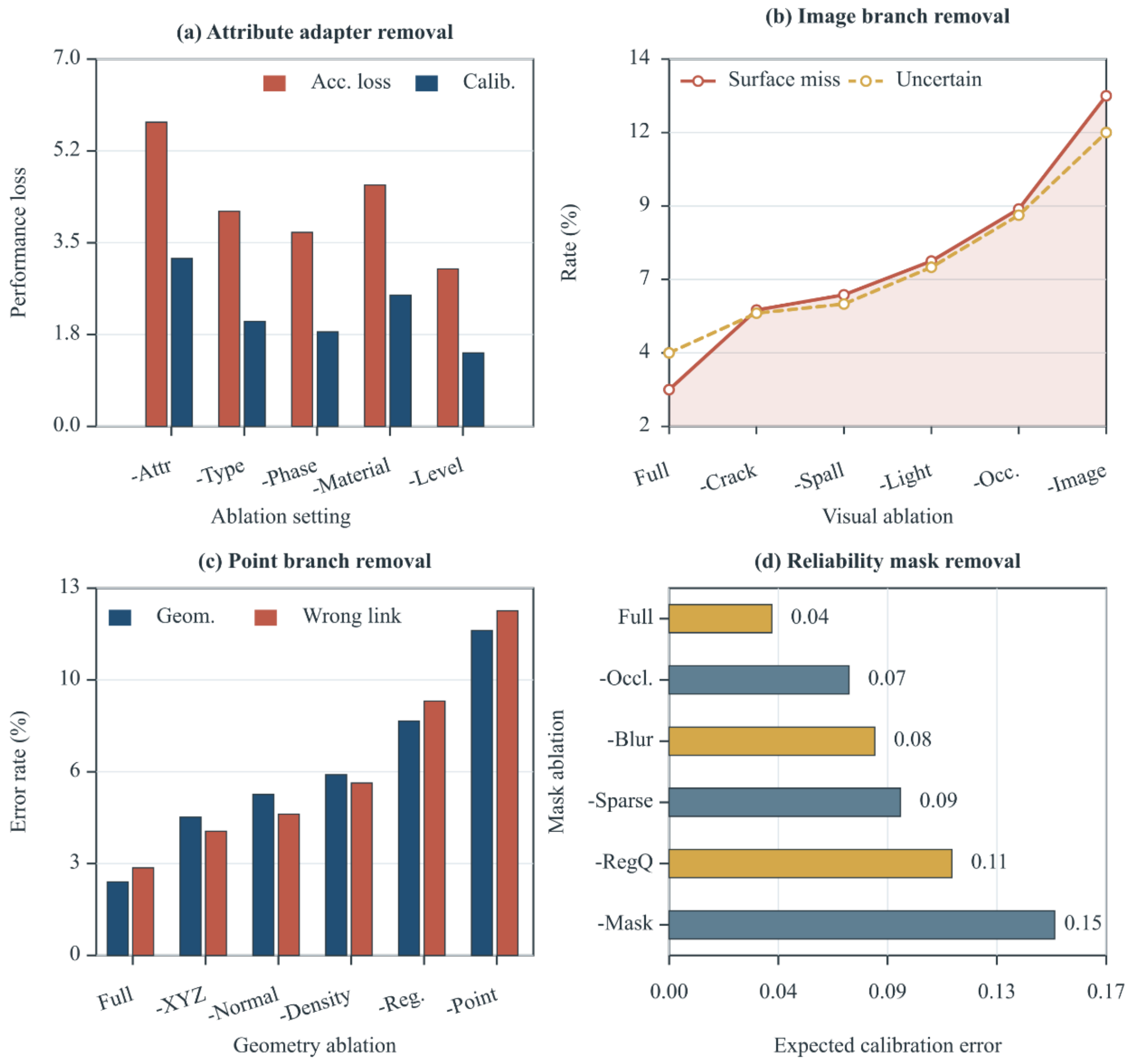

**Figure 6. Attribute-adapter and source-ablation analysis. (a) attribute adapter removal. (b) image branch removal. (c) point branch removal. (d) reliability mask removal.**

### 4.3. Occlusion Robustness and Engineering-review Output

Robustness is tested with image occlusion, poor light, sparse point clouds, BIM registration error, repeated component layout and missing attribute data. These disturbances are common in construction inspection and should not all be treated as defects. Robust digital-inspection research shows that unseen site conditions affect model reliability [20], so the test set includes disturbance combinations outside training. Site-quality studies also show that repeated components and temporary shielding can create false associations [21], while confidence-aware multimodal fusion supports lowering certainty when evidence is weak [22].

Figure 7 is the robustness-test result. As shown in Figure 7(a), the image-occlusion panel lowers confidence when the visual branch cannot observe the defect region completely, while point and attribute evidence still support component association. As shown in Figure 7(b), the BIM-registration-error panel shows lower association accuracy when point-to-BIM alignment is poor, but the attribute adapter reduces this drop by using component context. Construction inspection requires reliable data sources [23], so low-confidence cases should be returned for review rather than closed automatically.

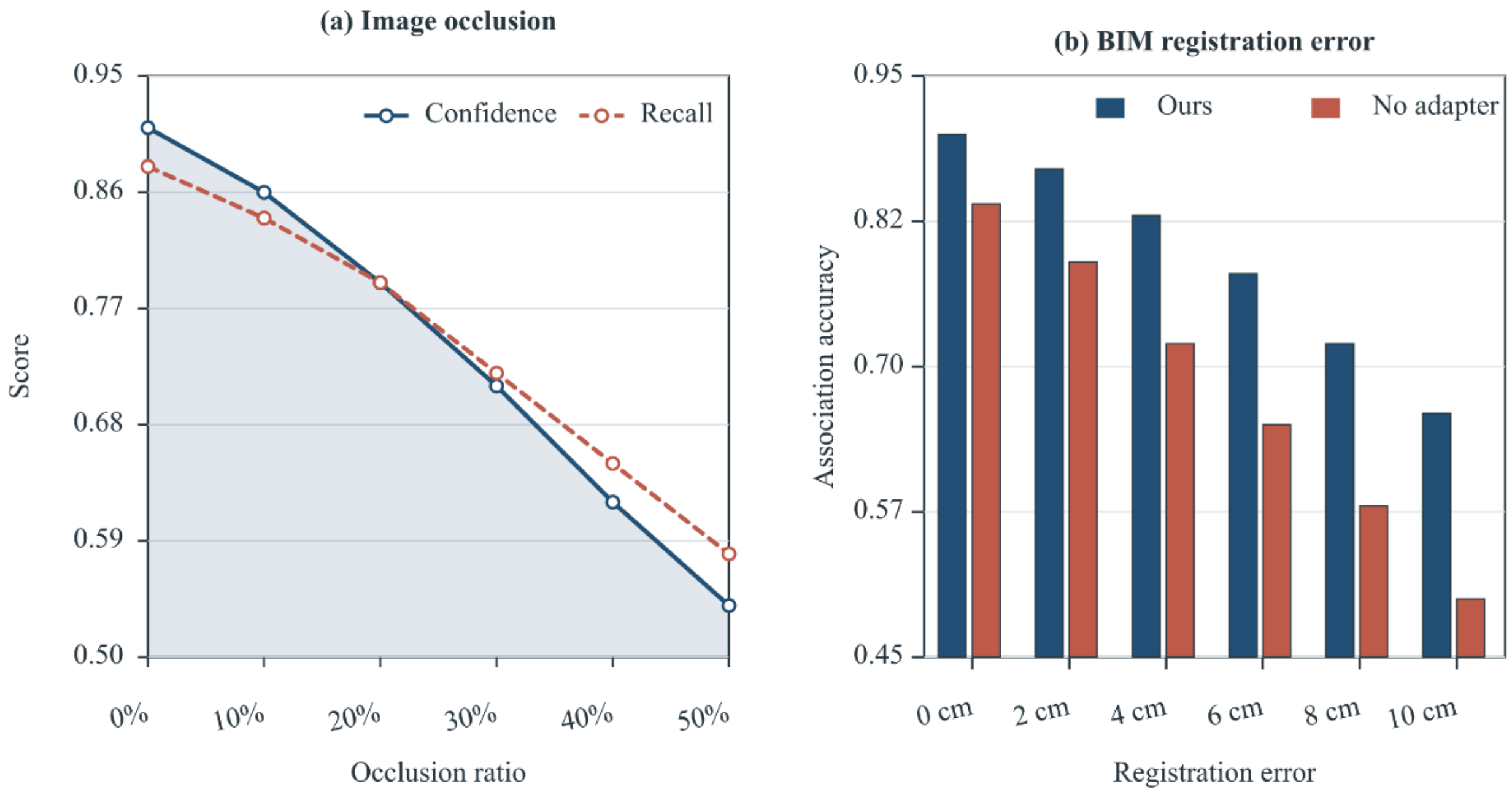

**Figure 7. Robustness under occlusion and registration disturbances. (a) image occlusion. (b) BIM registration error.**

The engineering-review output reports component link, defect probability, confidence, dominant evidence and recommended review action. If dominant evidence is visual, the engineer can revisit the site image or capture a clearer photo. If dominant evidence is point geometry, the team can check alignment and geometric deviation in the scan. If evidence is attribute-based, the engineer can confirm whether the BIM model and construction stage are current. Human-in-the-loop inspection studies support separating prediction from review through an audit trail [24], digital-twin issue management requires component identity and evidence type [25], and quality-control workflows should expose uncertainty before closing a defect [26].

Figure 8 is the review output of the proposed system. Figure 8(a) shows the review card containing component ID, defect class, defect probability, confidence and recommended action. Figure 8(b) shows dominant evidence, including image damage, point deviation, BIM attribute mismatch and source unreliability. Evidence-linked construction analytics emphasize traceability from observation to component [27], and component dashboards can group repeated defects by floor, zone, trade and source type [28]. The review card also preserves linkage reasons and confidence at inspection time [29, 30], so the two panels support issue assignment, re-scan decisions, new photographs and manual component inspection.

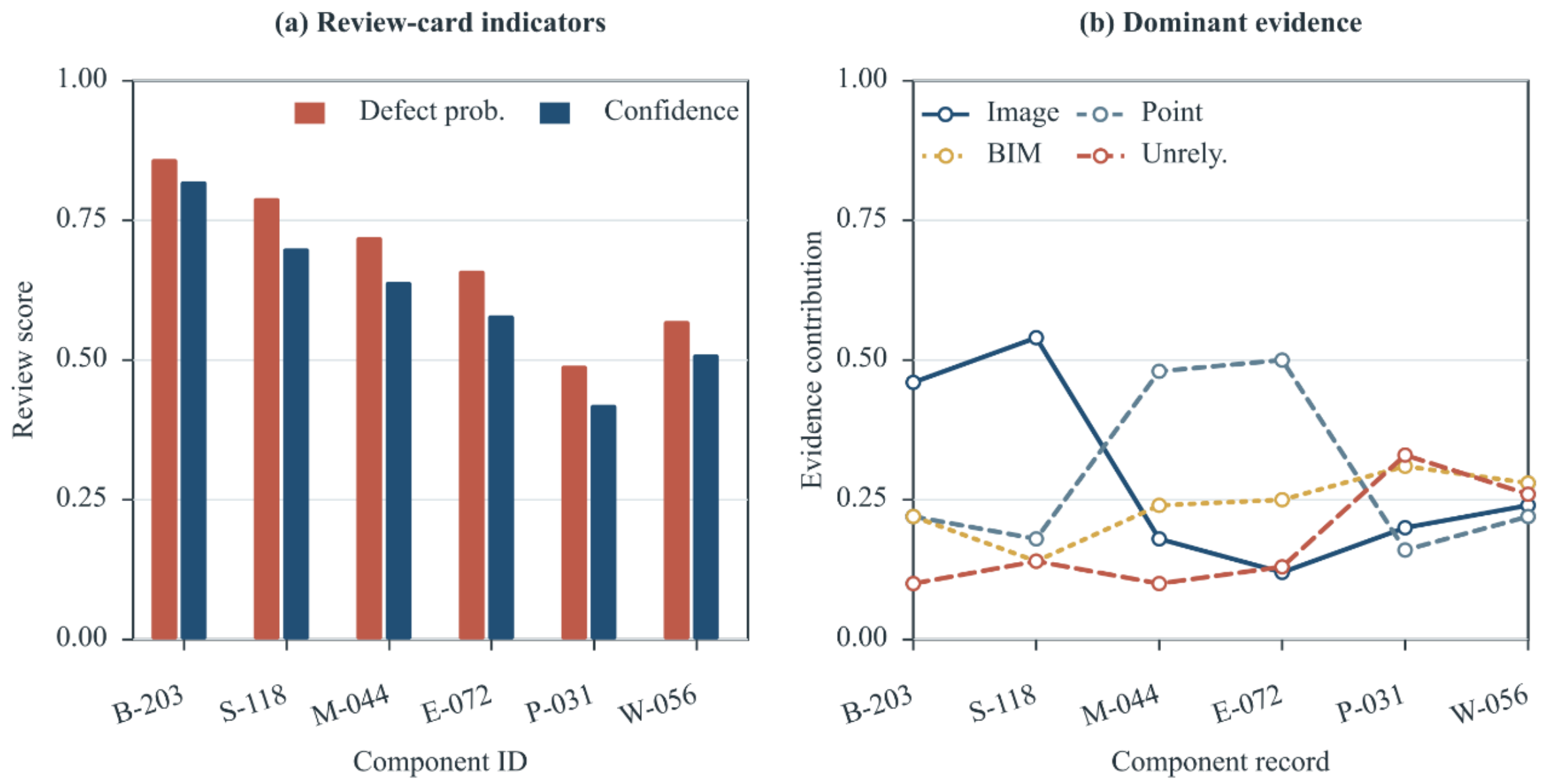

**Figure 8. BIM component defect review card and dominant-evidence output. (a) review card. (b) dominant evidence.**

Based on the error analysis, false associations are most likely in repeated areas where adjacent pipe supports, sleeves or embedded plates look similar. False defect alarms may also arise from temporary stains, and sparse point clouds may create artificial gaps. BIM-based quality-management studies show that these errors should be corrected through data governance and model coordination [31], while standard-aware inspection can connect predicted defects with tolerance libraries [32]. Long-term records should keep raw image crops, point segments and component IDs [33]. Image-point cloud fusion studies further show that registration and filtering influence defect mapping [34], and BIM component decision support benefits from explicit specification context [35]. The model should therefore be used to identify inspection weaknesses and support review, not to approve or close defects automatically.

## 5. Conclusion

This paper proposes a PointNeXt-Adapter-improved model for BIM component defect association based on construction images and design model attributes. Align the site image and local point cloud with BIM component attributes and source-reliability tags to build component-level inspection windows. PointNeXt learns local geometry and deviation, and an adapter adds design semantics to the point representation. The fusion head generally outputs component association, defect state, confidence, dominant evidence, and review action.

Based on the above experiments, the proposed model has improved association accuracy, defect recall and confidence calibration over the image-only, point-only, simple-fusion and adapter-free baseline methods. Based on the analysis of the data table and figures, component type, defect type, image quality, point density and matching quality all impact performance. Robustness tests have also shown that the model can reduce overconfidence caused by occlusion and registration errors and keep relevant evidence for BIM-based inspection.

Future work should test the framework with real construction-site data, richer BIM attributes and project-specific tolerance rules. The model can be connected to mobile scanning, image-capture planning, BIM issue-management platforms and digital twin dashboards. Deployment should also include BIM update checks, scan-density control, repeated-layout handling and expert confirmation for low-confidence cases. These extensions would support traceable quality-control workflows that link field evidence, design-model attributes and engineering review actions through a shared component-level representation.

## 6. Declarations

### 6.1. Author Contributions

Na Yu contributes to conceptualization, methodology, software, validation, analysis, investigation, data collection, draft preparation, manuscript editing, visualization, supervision, project administration, and funding acquisition. All authors have read and agreed with the manuscript before its submission and publication.

### 6.2. Funding and Acknowledgments

This research received no specific financial support from any funding agency.

### 6.3. Institutional Review Board Statement

Not applicable.

### 6.4. Informed Consent Statement

Not applicable.

### 6.5. Declaration of Competing Interest

Not applicable.

### 6.6. Declaration of Competing Interest

The authors declare that they have no known competing financial interests or personal relationships that could have appeared to influence the work reported in this paper.

### 6.7. Declaration of generative AI and AI-assisted technologies

During the preparation of this manuscript, the authors utilized Grammarly for linguistic polishing, logical organization and draft revision. The generative AI tool was not involved in research design, data collection, data analysis, or the equation of core conclusions. All outputs generated by AI were reviewed, revised and verified manually by the authors. The authors take full responsibility for all content of this manuscript.